\documentclass[11pt]{article}

\usepackage[preprint]{acl}

\usepackage{times}
\usepackage{latexsym}

\usepackage[T1]{fontenc}

\usepackage[utf8]{inputenc}

\usepackage{microtype}

\usepackage{inconsolata}

\usepackage{graphicx}

\usepackage{amsmath}
\usepackage{amssymb}
\usepackage{booktabs}
\usepackage{subcaption}
\usepackage{multirow}
\usepackage{tikz}

\title{Multilinguality as Sense Adaptation}

\author{
 Jan Christian Blaise Cruz\textsuperscript{1,5}\thanks{Work was done while at Mila - Quebec AI and McGill University.}\quad
 David Ifeoluwa Adelani\textsuperscript{2,3,4}\quad
 Alham Fikri Aji\textsuperscript{1,5}\\\\
 \textsuperscript{1}MBZUAI\quad
 \textsuperscript{2}McGill University\quad
 \textsuperscript{3}Mila - Quebec AI Institute\quad\\
 \textsuperscript{4}Canada CIFAR AI Chair\quad
 \textsuperscript{5}SEACrowd\quad\\
 \texttt{\{jan.cruz,alham.fikri\}@mbzuai.ac.ae} \quad \texttt{david.adelani@mila.quebec}
}

\begin{document}
\maketitle

\begin{abstract}
We approach multilinguality as \emph{sense adaptation}: aligning latent meaning representations across languages rather than relying solely on shared parameters and scale. 
In this paper, we introduce \textbf{SENse-based Symmetric Interlingual Alignment} (SENSIA), which adapts a Backpack language model from one language to another by explicitly aligning sense-level mixtures and contextual representations on parallel data, while jointly training a target-language language modeling loss to preserve fluency. 
Across benchmarks on four typologically diverse languages, \textsc{SENSIA} generally outperforms comparable multilingual alignment methods and achieves competitive accuracy against monolingual from-scratch baselines while using 2–4$\times$ less target-language data. 
Analyses of learned \textit{sense geometry} indicate that local sense topology and global structure relative to English are largely preserved, and ablations show that the method is robust in terms of design and scale.
Our results show that sense adaptation provides a simple, data-efficient, and linguistics-motivated path to multilingual fluency that is useful to low-resource languages and motivates further linguistics-motivated research.
\end{abstract}

\section{Introduction}

Multilingual language models now underpin cross-lingual NLP, from encoder-only systems such as XLM-R \cite{conneau2020xlmr} to generative models like mT5 \cite{xue-etal-2021-mt5}, XGLM \cite{lin-etal-2021-xglm}, and BLOOMZ \cite{muennighoff2022bloomz}.
However, despite strong empirical gains, we still lack a clear understanding of how cross-lingual competence arises: alignment often emerges only implicitly, can be uneven across layers, and is sometimes brittle across languages and tasks~\citep{hammerl-etal-2024-understanding,wang-etal-2024-probing-emergence,liu2025midalign,li-etal-2024-improving-context}.
Many existing approaches are largely engineering-driven: they often derive
performance from parameter sharing and increasing data or compute, without
explicitly specifying what units of meaning should be aligned.


In this work, we adopt a linguistically motivated view: a model is multilingual not just because it processes several languages in a shared parameter space, but because it represents similar meanings with comparable internal ‘senses’ across languages.
Building on this view, we propose \textbf{SENse-based Symmetric Interlingual Alignment} (\textsc{SENSIA}), a method for aligning sense‑level representations across languages.
Concretely, \textsc{SENSIA} starts from an English Backpack \cite{hewitt2023backpack} language model, which represents each word as a mixture of latent sense vectors. It adapts this model to a target language by aligning sense mixtures and contextual states with a symmetric InfoNCE objective \cite{oord2018representation} on parallel data, while jointly training a target-side LM loss to preserve fluency.


\begin{figure*}[t]
\centering
\includegraphics[width=\textwidth]{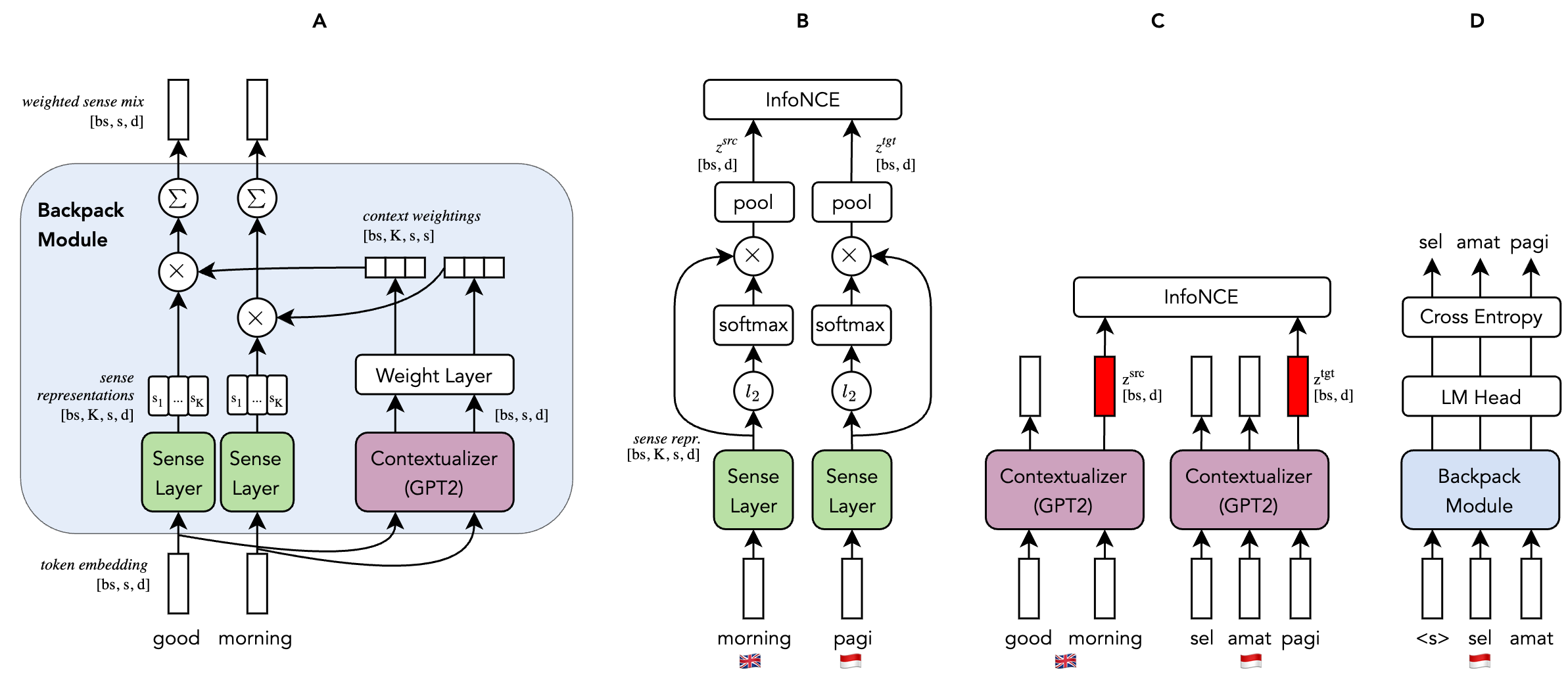}
\caption{\textbf{\textsc{SENSIA} overview.}
(A) A Backpack represents each input token as a context-dependent mixture of $K$ latent sense vectors.
(B) The \emph{sense loss} aligns sense-level representations of meaning.
(C) The \emph{context loss} aligns context-level representations via the last non-pad tokens.
(D) A target-language \emph{language modeling loss} preserves fluency while senses and contexts are aligned across languages.}
\label{fig:banner-image}
\vspace{-0.5em}
\end{figure*}

Across benchmarks on four typologically diverse languages, \textsc{SENSIA}‑adapted models outperform comparable dense‑representation multilingual alignment methods. They are also competitive with monolingual from‑scratch baselines in the same parameter regime while using 2 to 4$\times$ less target‑language data.
Structural ablations show that the original English local and global sense representations remain largely intact, providing evidence that \textsc{SENSIA} operates at the level of meaning.
Scaling tests show consistent improvements for all languages and tasks as model size increases. Additional empirical ablations indicate that our training recipe is robust, with each component contributing to final performance and fluency. Finally, we find that \textsc{SENSIA} degrades gracefully in low‑data regimes, making it attractive for low‑resource languages where curated monolingual data is scarce for large‑scale pretraining.


Our contributions are as follows:
\begin{itemize}
    \item We \textbf{model multilinguality as sense adaptation}, grounding cross‑lingual transfer in representation‑level alignment of meaning rather than in parameter sharing alone.
    \item We propose a \textbf{data‑efficient} sense‑based alignment method that adapts a pretrained model from one language to another using only parallel data, avoiding retraining from scratch.
    \item We demonstrate \textbf{stronger performance} than dense‑representation alignment and competitive accuracy with monolingual from‑scratch baselines, together with empirical evidence of meaning preservation, design robustness, and consistent performance at scale.
\end{itemize}

Ultimately, we argue that multilinguality can be modeled \textit{as} sense adaptation, and that \textsc{SENSIA} is a simple, robust, data-efficient, and linguistically-motivated operationalization of this principle.
We hope this reframing encourages more linguistically motivated work on \emph{what} to align (senses vs.\ parameters) and \emph{where} to align (layers/subspaces), yielding more reliable and inclusive cross-lingual transfer.

\section{Methodology}
\label{sec:methodology}

\paragraph{Overview.}
\textsc{SENSIA} uses the Backpack architecture \cite{hewitt2023backpack} as a base for cross-lingual adaptation.
Backpack augments a GPT-2 transformer with a sense layer that represents each input token $t$ as a mixture of $K$ latent sense vectors $\mathbf{s}_{t,1},\dots,\mathbf{s}_{t,K}$.
\footnote{In Backpack models, ``senses'' are not discrete, interpretable categories. Instead, they are continuous latent representations that capture possible variations in meaning. They provide a structured approximation of sense distinctions rather than explicit symbolic senses.}
Given the token embedding $\mathbf{x}_t$, the sense module produces these $K$ vectors, and an attention-based sense-weight layer assigns a distribution of importance weights $\pi_{t,k}$ over senses (and positions) conditioned on context.
The resulting context-dependent token representation is the soft mixture $\mathbf{h}_t = \sum_{k=1}^K \pi_{t,k}\,\mathbf{s}_{t,k},$
which replaces the standard token embedding and is consumed by the language-modeling head.
Aligning these soft sense mixtures and their contextualized representations across languages allows \textsc{SENSIA} to transfer conceptual structure while permitting language-specific reorganization of the sense space.

\subsection{Training Objectives}
\label{sec:training-objectives}

We perform contrastive learning on parallel sentence pairs using three complementary objectives: a \textbf{sense loss}, a \textbf{context loss}, and a \textbf{language modeling loss}.
A visual sketch of how these are computed can be found in Figure~\ref{fig:banner-image}.

\paragraph{Symmetric InfoNCE.}
Given a batch of $B$ normalized sentence-level vectors from the source language $\{\mathbf{u}_i^{\text{src}}\}_{i=1}^B$ and target language $\{\mathbf{u}_i^{\text{tgt}}\}_{i=1}^B$, and a temperature $\tau$, we define bidirectional InfoNCE probabilities
\begin{align}
p_{ij}(\tau)
&=
\frac{
  \exp\!\big(\mathbf{u}_i^{\text{src}} \cdot \mathbf{u}_j^{\text{tgt}} / \tau\big)
}{
  \sum_{k=1}^B \exp\!\big(\mathbf{u}_i^{\text{src}} \cdot \mathbf{u}_k^{\text{tgt}} / \tau\big)
},
\label{eq:infonce-p}\\[0.25em]
p'_{ij}(\tau)
&=
\frac{
  \exp\!\big(\mathbf{u}_i^{\text{tgt}} \cdot \mathbf{u}_j^{\text{src}} / \tau\big)
}{
  \sum_{k=1}^B \exp\!\big(\mathbf{u}_i^{\text{tgt}} \cdot \mathbf{u}_k^{\text{src}} / \tau\big)
}.
\label{eq:infonce-pprime}
\end{align}
The associated symmetric InfoNCE loss is
\begin{equation}
\begin{split}
\label{eq:infonce}
L_{\text{InfoNCE}}(\mathbf{u}^{\text{src}},\mathbf{u}^{\text{tgt}};\tau)\\
= -\frac{1}{2B} \sum_{i=1}^B \big[\log p_{ii}(\tau) + \log p'_{ii}(\tau)\big]
\end{split}
\end{equation}
where $p_{ii}(\tau)$ and $p'_{ii}(\tau)$ denote the probabilities assigned to the true translation pair (the $i$-th source and target sentences).

\paragraph{Sense loss.}
This objective encourages semantically equivalent words across languages to produce similar mixtures of sense embeddings.
Given a token embedding $\mathbf{x}_t$, the model computes $K$ sense vectors
\[
\mathbf{s}_{t,k} = \mathbf{E}_k^{\top}\mathbf{x}_t,
\qquad
k=1,\dots,K,
\]
where $\mathbf{E}_k$ are learned projection matrices that map token embeddings into distinct sense subspaces.
The magnitude of each sense vector, $a_{t,k} = \|\mathbf{s}_{t,k}\|_2$, serves as a measure of its contextual relevance and is transformed into a soft weighting distribution $\pi_{t,k} = \mathrm{softmax}_k\!\big(a_{t,k} / \tau_{\text{pool}}\big),$ where $\tau_{\text{pool}}$ is a temperature parameter controlling the sharpness of the weighting.
After computing $\mathbf{h}_t$ through the backpack, we obtain sentence-level sense embeddings $\mathbf{z}_i^{\text{sns}}$ by mean-pooling $\mathbf{h}_t$ over all non-padding tokens in sentence $i$ and $\ell_2$-normalizing.

For normalized source and target sense embeddings $\{\mathbf{z}_i^{\text{src},\text{sns}}, \mathbf{z}_i^{\text{tgt},\text{sns}}\}_{i=1}^B$, the sense alignment loss instantiates Eq.~\eqref{eq:infonce} with temperature $\tau_{\text{sns}}$:
\[
L_{\text{sns}} = L_{\text{InfoNCE}}\big(\mathbf{z}^{\text{src},\text{sns}}, \mathbf{z}^{\text{tgt},\text{sns}}; \tau_{\text{sns}}\big).
\]

\paragraph{Context loss.}
This objective aligns contextualized transformer representations, encouraging words with similar meanings to appear in comparable linguistic environments.
For each sentence $i$, we take $\mathbf{z}_i^{\text{ctx}}$ to be the final hidden state of the last non-padding token (again $\ell_2$-normalized), computed separately for source and target.
Using the same symmetric InfoNCE formulation in Eq.~\eqref{eq:infonce} with temperature $\tau_{\text{ctx}}$, we define
\[
L_{\text{ctx}} = L_{\text{InfoNCE}}\big(\mathbf{z}^{\text{src},\text{ctx}}, \mathbf{z}^{\text{tgt},\text{ctx}}; \tau_{\text{ctx}}\big).
\]

\paragraph{Language modeling loss.}
To improve fluency in the target language, the model minimizes a label-smoothed cross-entropy loss over target tokens:
\[
L_{\text{lm}} = -\sum_t q_t \log p_\theta\big(y_t \mid x^{\text{tgt}}_{<t}\big),
\]
where $y_t$ is the target token at position $t$, $p_\theta(\cdot \mid x^{\text{tgt}}_{<t})$ is the model’s conditional distribution, and $q_t = (1-\varepsilon)\,\delta_{y_t} + \frac{\varepsilon}{|\mathcal{V}|}$ is the smoothed target distribution with vocabulary $\mathcal{V}$ and smoothing parameter $\varepsilon$.

\paragraph{Total objective.}
The overall training loss combines all components:
\[
L_{\text{total}} = w_{\text{sns}} L_{\text{sns}} + w_{\text{ctx}} L_{\text{ctx}} + w_{\text{lm}} L_{\text{lm}}.
\]
Training proceeds in three phases with distinct weight schedules:
an \emph{alignment} phase with high $w_{\text{sns}}$ and $w_{\text{ctx}}$ and negligible $w_{\text{lm}}$ to establish cross-lingual correspondence;
a \emph{joint} phase in which $w_{\text{lm}}$ is gradually increased while $w_{\text{sns}}$ and $w_{\text{ctx}}$ are reduced, allowing target-language fluency to develop while preserving semantic alignment; and
a \emph{polish} phase where $w_{\text{lm}}$ dominates but small alignment weights are retained as regularizers.
During the polish phase, the sense vectors and the sense-weight network are frozen to prevent representational drift, while the transformer backbone and language-modeling head continue to update.
Loss weights are smoothly annealed across phases using the cosine schedule described in Appendix \ref{app:scheduler}.

\subsection{Experimental Setup}
\label{sec:adaptation-setup}

\begin{figure*}[t]
\centering
\begin{subfigure}[t]{0.32\textwidth}
    \centering
    \includegraphics[width=\linewidth]{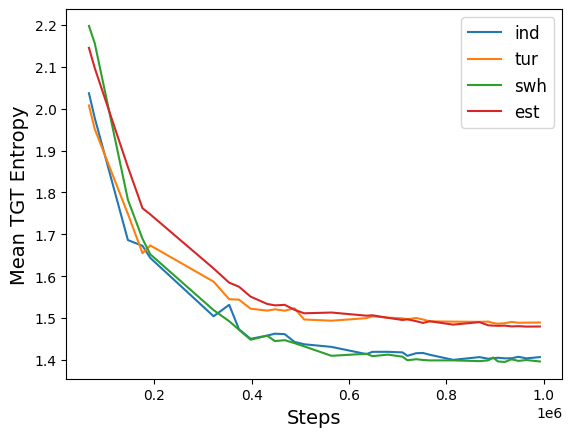}
    \caption{Mean target-side sense entropy.}
    \label{fig:entropy}
\end{subfigure}
\hfill
\begin{subfigure}[t]{0.32\textwidth}
    \centering
    \includegraphics[width=\linewidth]{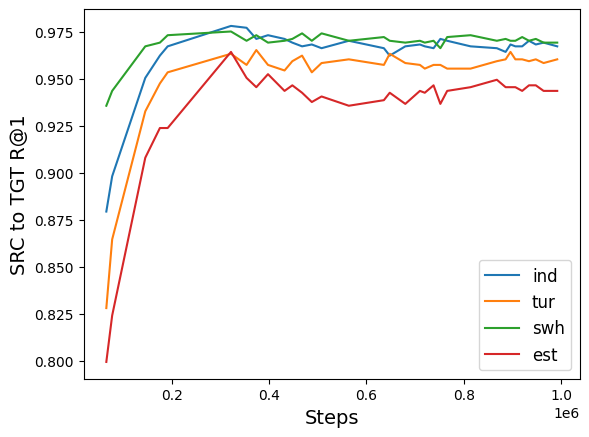}
    \caption{Source$\rightarrow$target R@1.}
    \label{fig:r1_src_tgt}
\end{subfigure}
\hfill
\begin{subfigure}[t]{0.32\textwidth}
    \centering
    \includegraphics[width=\linewidth]{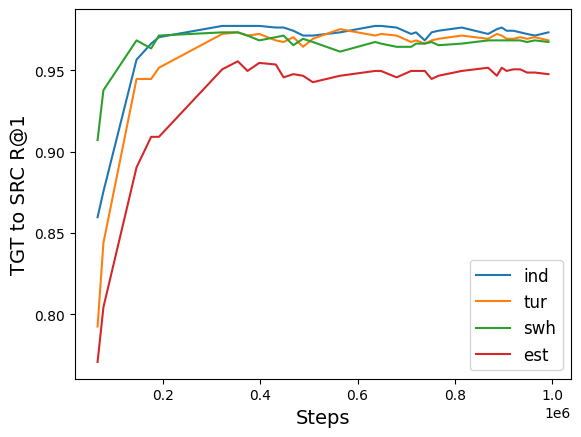}
    \caption{Target$\rightarrow$source R@1.}
    \label{fig:r1_tgt_src}
\end{subfigure}
\caption{\textbf{Training dynamics on the FLORES-200 validation set.} (a) Mean target-side sense entropy decreases and stabilizes as senses specialize without collapsing. (b--c) \texttt{src}$\rightarrow$\texttt{tgt} and \texttt{tgt}$\rightarrow$\texttt{src} recall@1 rapidly rise and remain high, indicating a stable, aligned bilingual space during adaptation.}
\label{fig:training_dynamics}
\vspace{-0.5em}
\end{figure*}

\begin{table*}[t]
\centering
\small
\setlength{\tabcolsep}{5pt}
\begin{tabular}{llcccccc}
\toprule
\textbf{Benchmark} & \textbf{Lang} & \textbf{XGLM-7B5} & \textbf{BLOOMZ-7B1} & \textbf{Goldfish} & \textbf{GPT-2 + FT} & \textbf{MCL} & \textbf{\textsc{SENSIA} (Ours)}  \\
\midrule
\multirow{4}{*}{\textbf{XCOPA}} 
 & Estonian   & 0.5700 & 0.4800 & 0.5200 & 0.4760 & 0.4760 & 0.5120 \\
 & Indonesian & 0.5800 & 0.6700 & 0.5840 & 0.4840 & 0.5120 & 0.5160 \\
 & Swahili    & 0.5700 & 0.6000 & 0.5500 & 0.5280 & 0.5120 & 0.5520 \\
 & Turkish    & 0.5600 & 0.4800 & 0.5620 & 0.5100 & 0.5140 & 0.5540 \\
\midrule
\multirow{2}{*}{\textbf{XStoryCloze}} 
 & Indonesian & 0.5963 & 0.8382 & 0.5321 & 0.5076 & 0.5162 & 0.5288 \\
 & Swahili    & 0.5533 & 0.6843 & 0.5321 & 0.5157 & 0.5407 & 0.5420 \\
\midrule
\multirow{4}{*}{\textbf{Belebele}} 
 & Estonian   & 0.3411 & 0.2988 & 0.2933 & 0.2522 & 0.2667 & 0.2911 \\
 & Indonesian & 0.3522 & 0.4744 & 0.2800 & 0.2344 & 0.2789 & 0.2911 \\
 & Swahili    & 0.3244 & 0.4022 & 0.2711 & 0.2489 & 0.2678 & 0.2844 \\
 & Turkish    & 0.3522 & 0.3311 & 0.3089 & 0.2656 & 0.2589 & 0.3067 \\
\bottomrule
\end{tabular}
\caption{\textbf{Accuracy on reasoning and comprehension benchmarks.} \textsc{SENSIA} is competitive against from-scratch training (Goldfish), outperforms both finetuning (GPT-2 + FT) and dense-representation multilingual alignment (MCL), and approaches the performance of much larger multilingual models.}
\label{tab:reasoning_comprehension}
\end{table*}

We adapt a pretrained English Backpack model to a target language using parallel data from OPUS \cite{tiedemann2012opus}. Adaptation is evaluated on four typologically diverse languages---Estonian, Turkish, Indonesian, and Swahili---selected for their linguistic variety and coverage in our benchmark datasets. 
These languages span distinct families (Finno-Ugric, Turkic, Austronesian, and Bantu) and vary widely in morphology and syntax, offering a strong test of whether sense-based alignment generalizes beyond Indo-European structure. Evaluation is then done on two angles: semantic alignment and task performance.

\paragraph{Semantic Alignment.} We evaluate on FLORES-200 during training, measuring bidirectional recall@1 for bilingual retrieval and the mean entropy of the target-side sense weights. Entropy characterizes the specialization of the sense inventory: values near $\log_{e} K$ indicate diffuse, non-specialized senses, while values near zero suggest sense collapse. These metrics capture how effectively the model forms a stable, aligned bilingual space. We additionally track perplexity on the FLORES-200 devtest set as an internal metric to gauge polish phase effectiveness\footnote{Perplexity is not compared against other models as different models use different tokenizers, and thus will not provide an apples-to-apples comparison.}. 

\paragraph{Task Performance.} Downstream performance is evaluated on XCOPA, XStoryCloze, and Belebele to probe reasoning and understanding.

\paragraph{Baselines.} We employ three main baselines: Goldfish~\citep{chang2024goldfish}: a suite of monolingual GPT-2 models trained \emph{from-scratch} with 1 GB of data; English GPT-2 \emph{finetuned} with the same target-language data and number of tokens as the \textsc{SENSIA}-adapted models, and; Multilingual Contrastive Learning (MCL) \cite{li-etal-2024-improving-context}\footnote{
We implement only the MCL phase of the AFP framework proposed by \citet{li-etal-2024-improving-context}. We do not include the second phase, Cross-Lingual Instruction Following, because our models are not instruction-following LLMs. Accordingly, we refer to this baseline as MCL rather than AFP.}, which adapts a model by aligning transformer representations with parallel data. This last method is the closest to ours and represents a baseline difference in \textit{alignment target}: MCL wants to align \textit{dense} transformer representations, while \textsc{SENSIA} wants to align \textit{meaning} representations through senses.
We also compare to larger multilingual models: XGLM-7B5~\citep{lin-etal-2021-xglm} and BLOOMZ-7B1~\citep{muennighoff2022bloomz}, which serve as upper bounds on multilingual transfer. Hyperparameter, data-processing, and benchmarking details appear in Appendix \ref{app:adaptation-setup}, \ref{app:data-construction}, and \ref{app:benchmarking}.

\section{Results and Discussion}
\label{sec:results}

\paragraph{\textsc{SENSIA} is stable and robust.}
During adaptation, mean target-side sense entropy decreases rapidly and stabilizes around 1.4--1.5, indicating specialization without collapse; bidirectional recall@1 on FLORES-200 converges near 0.97 for most languages, evidencing a stable bilingual mapping (Figure \ref{fig:training_dynamics}). Swahili and Indonesian reach this plateau fastest, while Estonian converges more slowly. These intrinsic signals align with the steady downstream behavior in Table~\ref{tab:reasoning_comprehension}.

\paragraph{\textsc{SENSIA} is competitive with from-scratch training, consistently beats finetuning, and narrows the gap to multilingual LMs.}
Against GPT-2 + FT, \textsc{SENSIA} is uniformly stronger across XCOPA, XStoryCloze, and Belebele (Table~\ref{tab:reasoning_comprehension}), with gains of +2.1 to +5.7~pp (mean +3.56~pp).
Relative to Goldfish, \textsc{SENSIA} is within $\pm$1~pp on 7/10 cells and ahead on 4 (XCOPA-Swahili +0.2~pp; XStoryCloze-Swahili +0.99~pp; Belebele-Indonesian +1.11~pp; Belebele-Swahili +1.33~pp), with a single notable deficit on XCOPA-Indonesian (--6.8~pp).
While absolute scores trail larger multilingual LMs, \textsc{SENSIA} narrows the gap in some settings (e.g., XCOPA-Turkish is within 0.6~pp of XGLM-7B5), highlighting the efficiency of sense alignment relative to broad multilingual pretraining.

\begin{table}[t]
\centering
\begin{tabular}{lccc}
\toprule
\textbf{Lang}  & \textbf{Total Tok.} & \textbf{Data (MB)} & \textbf{vs. GF} \\
\midrule
Estonian   & 83M  & 246 & 3.9$\times$ \\
Indonesian & 149M & 509 & 2.0$\times$ \\
Swahili    & 135M & 426 & 2.3$\times$ \\
Turkish    & 98M  & 283 & 3.5$\times$ \\
\bottomrule
\end{tabular}
\caption{\textbf{Training data efficiency} of \textsc{SENSIA} relative to Goldfish-1GB (GF). \textsc{SENSIA} achieves comparable performance with $2$--$4\times$ less target language data. Tokens are counted based on English GPT-2 tokenizer. MB is approximated based on UTF-8 bytes per token.}
\label{tab:data_efficiency}
\end{table}

\paragraph{Sense-based adaptation is stronger than dense-representation alignment.}
Across the 10 benchmark--language combinations where it is evaluated, MCL improves over GPT-2 + FT by roughly $+1.2$~pp on average, and is particularly helpful on XStoryCloze and Belebele.
However, \textsc{SENSIA} consistently matches or outperforms MCL in all cells, with an average gain of about $+2.3$~pp.
The largest improvements appear on XCOPA--Swahili and XCOPA--Turkish (up to $+4$--$5$~pp), and on Belebele--Turkish ($+4.8$~pp).
In a few settings (e.g., XCOPA--Indonesian and XStoryCloze--Swahili), the gap nearly closes ($\leq$0.4~pp), suggesting that LM-level contrastive alignment can recover part of the benefit, but sense-based alignment never underperforms it and typically yields a stronger model.

\paragraph{\textsc{SENSIA} is more data efficient.}
\textsc{SENSIA} achieves competitive accuracy with substantially smaller target-language budgets than Goldfish: $2.0\times$ to $3.9\times$ less data across languages (Table~\ref{tab:data_efficiency}), while maintaining parity or better on many rows of Table~\ref{tab:reasoning_comprehension}. Taken together, these results show that sense alignment preserves downstream quality while reducing the target-language data requirement by roughly $2$--$4\times$.

\section{Sense Ablations}
\label{sec:sense-ablations}

\paragraph{Is sense geometry truly preserved?} We ask whether SENSIA truly adapts senses or merely aligns words. Direct word-to-word comparison is unreliable under the shared English
BPE tokenizer, where a single target-language word can span many
subwords. Instead, we test whether the English sense space is preserved after
adaptation, both locally (within a word) and globally (across the
vocabulary). 

We therefore conduct two complementary analyses: \emph{Sense Topology analysis}, which measures whether each word's local sense cluster maintains its internal geometry, and \emph{Procrustes Analysis}, which tests whether a single orthogonal transformation can align the global target sense manifold to English. Together, these analyses probe whether adaptation reorganizes meaning rather than memorizing lexical correspondences. Theoretical and practical implementation details, including sampling and statistics, are given in Appendix \ref{app:extra-ablations}.

\begin{figure}[t]
\centering
\includegraphics[width=1.0\linewidth]{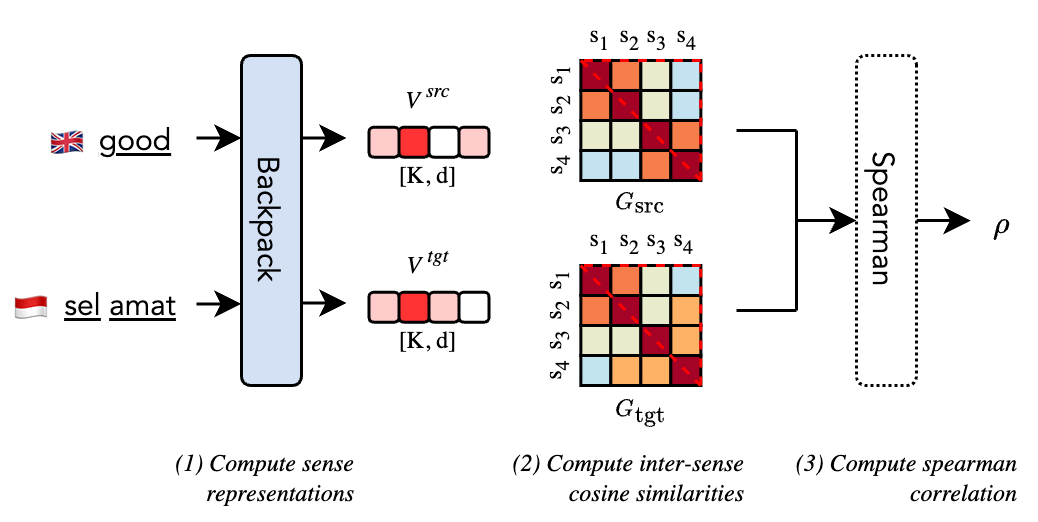}
\caption{\textbf{Sense Topology.} For each aligned word pair, we ask if its senses relate to each other similarly across languages, rather than requiring one-to-one matches between individual senses. We compute $K$ sense vectors for English and the target language, pooling if a word is split into subwords. We then build cosine gram matrices over senses in each language and compare their upper triangles with a Spearman correlation $\rho$. A high $\rho$ means that the pattern of inter-sense relationships is preserved even if the exact locations of individual senses differ, whereas exact one-to-one matches  indicates simple memorization of the English space.}
\label{fig:topology}
\vspace{-0.5em}
\end{figure}

\paragraph{Sense Topology Analysis.} (Figure \ref{fig:topology}) For each aligned English--target word pair, we ask whether its senses are arranged in a similar way across languages. Rather than requiring a one-to-one correspondence between individual senses, we focus on the \textit{pattern} of which senses are relatively close to or far from each other. Concretely, we compare the pairwise cosine-similarity structure of the $K$ sense vectors in each language: we construct cosine Gram matrices for English and the target language, then compute the Spearman correlation $\rho$ between their upper-triangular entries. Moderate $\rho$ values indicate that the local sense geometry is preserved but not trivially copied: extremely high values would suggest that the model is effectively memorizing the English configuration, whereas very low values would indicate substantial semantic drift.

\begin{figure}[t]
\centering
\includegraphics[width=0.9\linewidth]{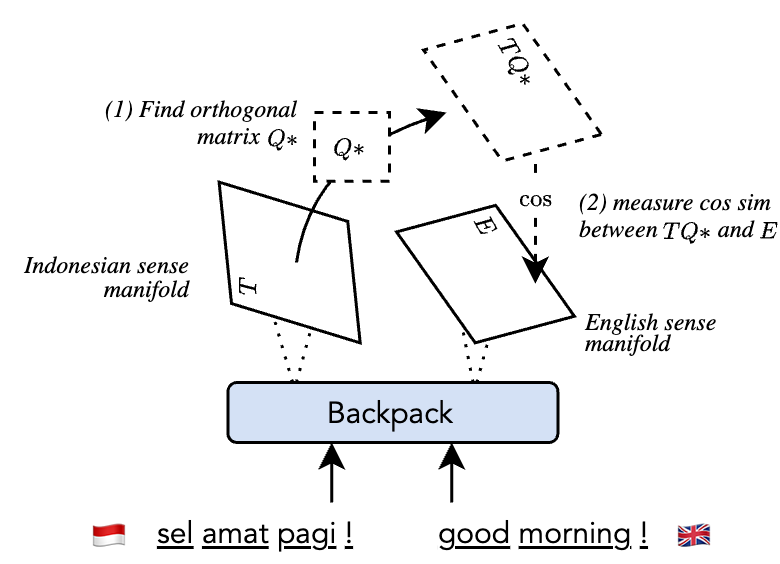}
\caption{\textbf{Procrustes Analysis. }Target-language sense embeddings may be rotated relative to English even if their structure is identical. We therefore learn an orthogonal transform $Q^{\star}$ that best aligns $T$ to $E$, and measure cosine similarity between the aligned spaces to assess how similar their global geometry is.}
\label{fig:procrustes}
\vspace{-0.5em}
\end{figure}

\paragraph{Procrustes Analysis.} (Figure \ref{fig:procrustes}) To assess whether global sense organization is preserved, we view the English and target-language sense mixtures as two embedding spaces and ask whether one can be mapped onto the other by a rigid transformation such as a rotation. For each aligned word pair, we compute a mixture-weighted sense embedding and then learn an orthogonal matrix that best maps target-language embeddings onto English via an orthogonal Procrustes problem. We finally measure cosine similarity between the aligned embeddings. High post-alignment cosine indicates that, up to a rotation, the target manifold has essentially the same overall shape as the English sense space.

\subsection{Results}

\begin{table}[t]
\centering
\small
\begin{tabular}{lccc}
\toprule
 & \textbf{\textsc{SENSIA} $\overline{\rho}$}  & \textbf{Control $\overline{\rho}$} & \textbf{$\Delta \rho$} \\
\midrule
est & 0.27 & 0.17 & \textbf{0.10} \\
ind & 0.30 & 0.21 & \textbf{0.09} \\
swh & 0.26 & 0.17 & \textbf{0.09} \\
tur & 0.25 & 0.16 & \textbf{0.09} \\
\bottomrule
\end{tabular}
\caption{\textbf{Sense Topology results.} Moderate correlation indicates that the model preserves local sense geometry relative to the control setup, showing that \textsc{SENSIA} keeps English meaning intact while accounting for language-specific variations.}
\label{tab:sense_topology_results}
\end{table}

\begin{table}[t]
\centering
\small
\begin{tabular}{lccc}
\toprule
 & \textbf{\textsc{SENSIA} $\overline{\cos}$} & \textbf{Control $\overline{\cos}$} & \textbf{$\Delta \cos$} \\
\midrule
est & 0.36 & 0.29 & \textbf{0.08} \\
ind & 0.44 & 0.33 & \textbf{0.11} \\
swh & 0.42 & 0.31 & \textbf{0.11} \\
tur & 0.35 & 0.26 & \textbf{0.09} \\
\bottomrule
\end{tabular}
\caption{\textbf{Procrustes Analysis results.} Moderate cosine similarity indicates that, up to an orthogonal mapping, the target space retains similar global structure to the original English sense space, showing that \textsc{SENSIA} successfully preserves learned English meaning at a global level.}
\label{tab:procrustes_results}
\end{table}

\paragraph{Sense-aware adaptation preserves both local and global structure.} As a control, we finetune an English Backpack to the target language using only the LM loss ($w_{\text{sns}} = w_{\text{ctx}} = 0$, $w_{\text{lm}} = 1.0$), which yields fluent target-language text but no explicit sense or context alignment.

Compared to this control, \textsc{SENSIA} consistently achieves higher local topology correlations ($\rho \approx 0.25$--$0.30$, $\Delta \rho \approx 0.09$--$0.10$; Table~\ref{tab:sense_topology_results}) and higher global Procrustes cosine similarities (0.35--0.44, $\Delta \cos \approx 0.08$--$0.11$; Table~\ref{tab:procrustes_results}).

Analytic languages such as Indonesian show slightly tighter alignment than agglutinative ones like Turkish, consistent with richer morphology. 
Overall, local sense clusters remain coherent, and the global manifold is well aligned with English up to an orthogonal mapping, supporting the view that \textsc{SENSIA} reorganizes meaning rather than destroying the original sense geometry.

\section{Empirical Ablations}
\label{sec:empirical-ablations}

\paragraph{Where does \textsc{SENSIA}'s performance come from?} We probe \textsc{SENSIA}'s design with four empirical ablations: 
(i) removing the sense-mixture mechanism,
(ii) ablating loss terms and training phases,
(iii) scaling the Backpack backbone, and
(iv) varying the amount of parallel data, plus a stress test on Chinese with an English tokenizer. These experiments pinpoint which components drive fluency and reasoning, and how robust the method is across scale and data regimes. Practical and implementation details can be found in Appendix \ref{app:extra-ablations}.

%
%

\begin{table}[t]
\centering
\small
\setlength{\tabcolsep}{4.5pt}
\begin{tabular}{lccc}
\toprule
\textbf{Language} & \textbf{Full CE} & \textbf{Top-1 CE} & \textbf{Uniform CE} \\
\midrule
Swahili    & \textbf{2.98} & 8.77 & 10.10 \\
Estonian   & \textbf{3.28} & 8.95 & 10.16 \\
Indonesian & \textbf{2.88} & 8.58 & 10.10 \\
Turkish    & \textbf{2.82} & 8.83 & 10.13 \\
\bottomrule
\end{tabular}
\caption{\textbf{Sense Mixture Ablation}. Cross-entropy on FLORES \textit{devtest} (lower is better). Results indicate that the model's performance relies on successful mixing and non-collapse of senses.}
\label{tab:sense_mixture_ablation}
\vspace{-0.5em}
\end{table}

\paragraph{Sense mixtures are necessary for fluent prediction.}
To test whether the model truly uses its sense mixtures at inference, we ablate the mixture weights $\pi_{t,k}$ while keeping all other parameters fixed. We compare three settings on FLORES devtest: the full learned mixture, a top-1 variant that renormalizes only the most probable sense per token, and a uniform mixture over all $K$ senses (Table~\ref{tab:sense_mixture_ablation}). Across four target languages, moving from the full mixture to top-1 increases token-level cross-entropy by roughly +6, and uniform mixtures degrade even further. This indicates that \textsc{SENSIA}'s fluency and predictive quality depend critically on softly combining multiple senses per token, and that this dependency is stable across typologically diverse languages.

%
%

\begin{table}[t]
\centering
\footnotesize
\setlength{\tabcolsep}{2.5pt}
\begin{tabular}{lcccc}
\toprule
\textbf{Variant} & \textbf{XCOPA} & \textbf{XStory} & \textbf{Belebele} & \textbf{PPL} \\
\midrule
No Context loss   & 0.498 & 0.523 & 0.278 & 13.83 \\
No Sense loss   & 0.484 & 0.524 & 0.252 & 13.64 \\
No LM loss    & 0.500 & 0.484 & 0.232 & 7.7e7 \\
No Align phase  & 0.512 & 0.453 & 0.258 & 13.66 \\
No Joint phase  & 0.498 & 0.456 & 0.278 & 14.57 \\
No Polish phase & 0.510 & 0.465 & 0.269 & 14.55 \\
\textbf{Full model} & \textbf{0.516} & \textbf{0.529} & \textbf{0.291} & \textbf{10.58} \\
\bottomrule
\end{tabular}
\caption{\textbf{Loss and phase ablations} (Indonesian target). Results indicate that all phases and losses play a role in the final performance of the adapted model.}
\label{tab:loss_phase_ablation}
\vspace{-0.5em}
\end{table}

\paragraph{Losses and training phases play complementary roles.}
We ablate each of the three objectives (Sense, Context, LM) and each phase in the Alignment--Joint--Polish schedule using Indonesian as the target language (Table~\ref{tab:loss_phase_ablation}). Removing the Context loss mostly hurts accuracy while leaving perplexity similar, showing that contextual alignment sharpens semantic decisions more than fluency. Dropping the Sense loss degrades comprehension benchmarks while leaving fluency relatively intact, indicating that sense-level anchoring primarily supports semantics. Eliminating the LM loss leads to extreme perplexity and lower accuracy, confirming that the generative objective is indispensable for stable training and coherent text. In the curriculum, skipping the Alignment phase causes moderate drops that can be partially recovered later, whereas removing the Joint phase yields the largest overall degradation, showing that jointly optimizing contrastive and generative signals is crucial. The Polish phase mainly refines fluency and phrasing, providing smaller but consistent gains.

%
%

\begin{table}[h]
\centering
\small
\begin{tabular}{llccc}
\toprule
\textbf{Task} & \textbf{Lang} & \textbf{Small} & \textbf{Medium} & \textbf{Large} \\
\midrule
XCOPA        & \textit{ind} & 0.4740 & 0.4840 & 0.5140 \\
                      & \textit{swh} & 0.5360 & 0.5440 & 0.5480 \\
XStoryCloze  & \textit{ind} & 0.5076 & 0.5156 & 0.5189 \\
                      & \textit{swh} & 0.5107 & 0.5281 & 0.5301 \\
Belebele     & \textit{ind} & 0.2611 & 0.2689 & 0.2733 \\
                      & \textit{swh} & 0.2567 & 0.2778 & 0.2911 \\
\midrule
FLORES  & \textit{ind} & 10.58 & 9.90 & 8.18 \\
                                   & \textit{swh} & 11.76 & 11.68 & 11.52 \\
\bottomrule
\end{tabular}
\caption{\textbf{Scaling test}. Results show predictable monotonic scaling, indicating that \textsc{SENSIA} benefits from scale.}
\label{tab:scaling-results}
\end{table}

\paragraph{Scaling improves reasoning and fluency in a predictable way.}
We pretrain Backpack models with GPT-2 small, medium (345M), and large (762M) backbones, following the same OpenWebText setup and Chinchilla-style token budgets \cite{hoffmann2022training}, then adapt each to Indonesian and Swahili under the same \textsc{SENSIA} recipe (details in Appendix \ref{app:extra-ablations}). Across both languages, moving from Small to Large increases macro-average accuracy over XCOPA, XStoryCloze, and Belebele by about +2.2 points and reduces FLORES perplexity (Table~\ref{tab:scaling-results}). These monotonic gains indicate that sense-based adaptation benefits from scale in a manner consistent with standard LM scaling laws.

%
%

\begin{table}[t]
    \centering
    \begin{tabular}{llcc}
        \toprule
        \textbf{Task} & \textbf{Lang} & \textbf{2M Pairs} & \textbf{100k Pairs} \\
        \midrule
        XCOPA       & ind & 0.5160 & 0.4880 \\
                    & swh & 0.5520 & 0.4940 \\
        XStoryCloze & ind & 0.5288 & 0.5122 \\
                    & swh & 0.5420 & 0.5107 \\
        Belebele    & ind & 0.2911 & 0.2600 \\
                    & swh & 0.2844 & 0.2733 \\
        \bottomrule
    \end{tabular}
    \caption{\textbf{Low-resource adaptation test.} Results show that the method is reasonably resistant to low-data regimes, and thus is useful for low-resource settings.}
    \label{tab:low-resource}
\end{table}

\paragraph{\textsc{SENSIA} degrades gracefully with 20$\times$ less bitext.}
To approximate low-resource conditions, we downsample the Indonesian and Swahili parallel corpora used for adaptation to 100k sentence pairs, keeping the architecture, schedule, and loss weights fixed and retraining from the same English initialization. Across XCOPA, XStoryCloze, and Belebele, accuracy drops by roughly 2.5 points for Indonesian and 3.3 points for Swahili (Table~\ref{tab:low-resource}), with macro accuracy decreasing by about 3 points overall despite a 20$\times$ reduction in bitext. This indicates that sense-based adaptation remains competitive in realistically low-data regimes.

%
%

\begin{table}[t]
    \centering
    \begin{tabular}{llcc}
        \toprule
        Task  & Goldfish & \textsc{SENSIA} (Zho) \\
        \midrule
        XCOPA       & 0.5420 & 0.4800 \\
        XStoryCloze & 0.4944 & 0.5069 \\
        Belebele    & 0.3144 & 0.2611 \\
        \bottomrule
    \end{tabular}
    \caption{\textbf{Hostile tokenizer test.} Results show the method's relative weakness when adapting to a language that uses non-Latin scripts.}
    \label{tab:chinese-test}
\end{table}

\paragraph{A hostile tokenizer limits sense adaptation on non-Latin scripts.}
We adapt the same English Backpack to Chinese using English--Chinese bitext from OPUS while keeping the original English GPT-2 BPE tokenizer unchanged, so Chinese is represented as long, poorly aligned subword sequences. Under this setup, \textsc{SENSIA} trails the monolingual Goldfish baseline by 6.2 points on XCOPA and 5.3 points on Belebele, while slightly surpassing Goldfish on XStoryCloze (+1.3 points; Table~\ref{tab:chinese-test}). Averaged across tasks, the adapted model remains 3--4 points behind Goldfish, suggesting that when tokenization is badly misaligned, sense adaptation cannot fully compensate. We therefore view this experiment as a stress test highlighting tokenizer limitations rather than evidence that \textsc{SENSIA} currently supports non-Latin scripts.

\section{Related Work}
\paragraph{LM-level cross-lingual alignment in causal LMs.}
A growing line of work targets the \emph{internal states} of decoder-only LMs to improve cross-lingual transfer. \citet{liu2025midalign} show that aligning \emph{middle layers} during task fine-tuning consistently boosts transfer, suggesting that layer choice is critical for multilingual sharing. Complementarily, \citet{li-etal-2024-improving-context} align \emph{sentence-level} hidden states in generative LMs via multilingual contrastive learning, while \citet{wang-etal-2024-probing-emergence} probe training checkpoints of BLOOM and relate the emergence of cross-lingual alignment to neuron overlap and downstream performance. 
\textbf{\textsc{SENSIA} differs in both \emph{target} and \emph{granularity}:} instead of aligning generic hidden states, we align a \emph{sense‑decomposed generator}—optimizing symmetric InfoNCE on \emph{latent sense mixtures} and a \emph{context} vector, while preserving target‑side fluency—building directly on Backpack’s sense factorization \citep{hewitt2023backpack}. This design is \emph{linguistically motivated}: the unit of transfer is meaning (senses), not parameters or subword overlap.

\paragraph{Activation‑space and post‑hoc interventions.}
Beyond gradient tuning, several works steer or map activation spaces to enhance multilinguality in LLMs. \citet{sundar-etal-2025-steering} show that neuron‑level \emph{model interventions} (“finding experts”) can measurably increase cross‑lingual alignment in multilingual LLMs; \citet{wang-etal-2025-bridging} propose \emph{INCLINE}, learning linear alignment matrices from parallel data and applying them at inference; and \citet{bandarkar2025layerswap} demonstrate \emph{layer swapping} as a post‑hoc merge that composes language and task experts for zero‑shot cross‑lingual transfer. Surveys synthesize broader evidence and cautions about what and where to align \citep{hammerl-etal-2024-understanding}. 
\textbf{In contrast,} we restructure the model’s \emph{semantic basis} by aligning \emph{latent senses}, then show that local topology and a near‑orthogonal global mapping are preserved after adaptation—empirical evidence for a \emph{linguistically motivated} view of multilinguality as \emph{sense adaptation} rather than surface‑form or parameter sharing.

\section{Conclusion}

We approach multilinguality as sense adaptation: aligning latent meaning representations across languages rather than merely sharing parameters. We operationalize this with \textsc{SENSIA}, which yields competitive performance against from-scratch, alignment, and finetuning methods. We additionally provide systemic ablations that show the method's soundness and robustness, as well as tests of scale and tokenizer limitations.
Looking forward, we aim to scale to larger backbones and explore ways to adapt more smoothly to non-Latin text. More broadly, we view sense adaptation as a simple, effective, and linguistically motivated path to multilingual fluency inside generative LMs.

\section*{Limitations}

While our reframing of multilinguality as \emph{sense adaptation} yields competitive accuracy and strong data efficiency, several limitations remain.

\paragraph{Tokenization and script coverage.}
Our pipeline assumes a shared GPT-2 BPE tokenizer trained on English and, in practice, targets Latin-script languages. All four evaluation languages (Estonian, Indonesian, Swahili, Turkish) use Latin script, and both data accounting and modeling rely on the English GPT-2 tokenizer (e.g., shared vocabulary of 50{,}264 tokens; data-efficiency counts based on the English tokenizer).\footnote{See Table~\ref{tab:scaling-results} for scaling results. Appendix \ref{app:base-configs} notes the shared vocabulary size.} This assumption restricts adaptation to languages that are not severely over-segmented by the English BPE; applying the method to non-Latin scripts (e.g., Chinese, Arabic, Devanagari) is understudied and out of scope in its current form. Future work will investigate adapting or retraining the tokenizer (e.g., Unicode/byte-level tokenization or lightweight target-specific merges) while preserving the learned sense geometry. 

\paragraph{Nature of supervision (OPUS) and scaling limits.}
Our adaptation is driven by sentence-parallel bitext filtered from OPUS. These pairs are typically short and literal with limited discourse continuity and world knowledge, which strengthens local semantic alignment and fluency but provides little grounding for discourse-level reasoning. In the scaling study (Table~\ref{tab:scaling-results}), \textsc{XStoryCloze} shows smaller Small$\to$Large gains than \textsc{XCOPA} on average (+1.5 vs.\ +2.6 points), and \textsc{Belebele} is intermediate (+2.3 points overall, with Swahili much stronger than Indonesian), consistent with the view that narrative plausibility and reading comprehension require information beyond sentence-level semantics. We hypothesize that OPUS’s length and domain characteristics cap improvements on these tasks. Future work will explore longer, document-level parallel corpora (e.g., paragraph/episode-aligned subtitles, newswire, TED talks) and grounded parallel sources to inject discourse and world knowledge.

\paragraph{Parallel-data dependence.}
\textsc{SENSIA} presumes adequate, high-quality alignments for the target language. This constraint may limit effectiveness in low-resource settings where bitext is scarce or noisy, or where domain mismatch is severe. Extending the approach to weakly aligned or mined corpora---with alignment uncertainty modeled in the loss---is a natural next step.

\paragraph{Model size regime and instruction-level transfer.}
Our experiments intentionally do not test scaling to billion-parameter models; the goal is to validate the representation-level framing, not to chase absolute SOTA. It remains an open question whether sense adaptation scales to 7B+ backbones and whether it can transfer task performance when supervision comes from a \emph{parallel instruction} corpus rather than sentence-parallel text. We leave both questions to future work.

\paragraph{On Chasing SOTA Results.}
We do \emph{not} claim state-of-the-art results or parity with large multilingual LLMs---and that is \textbf{not the objective of this paper}. Our aim is to introduce a linguistically motivated way of thinking about multilinguality, demonstrate that the reframing works (\textsc{SENSIA}), show it is competitive and data-efficient, and establish robustness through analyses and ablations (sense topology, Procrustes, mixture and loss/phase ablations). The next step is to evaluate the approach at LLM scale and test whether sense adaptation can transfer instruction-following and downstream tasks when trained with parallel instruction data.

\section*{Ethical Considerations}

\paragraph{Data.} Our adaptation relies on parallel corpora from OPUS, whose sub-corpora vary in license, domain, and quality; we apply length/similarity filtering but do not add PII scrubbing or content moderation. Because we align target languages to an English-trained model, English-centric biases and OPUS domain skews may carry over, especially for dialects or topics under-represented in bitext.

\paragraph{Risk.} Supervision is sentence-parallel and short-form, which chiefly improves \emph{semantic alignment and fluency} and less so discourse grounding; correspondingly, our fixed-protocol scaling gains are larger on \textsc{XCOPA} than on \textsc{XStoryCloze} or \textsc{Belebele} (+2.6 pp vs.\ +1.5 and +2.3 pp Small$\to$Large), consistent with those tasks requiring narrative plausibility and comprehension beyond sentence-level semantics (Table~\ref{tab:scaling-results}). We will investigate longer, document-level parallel corpora and grounded sources. Our scope is GPT-2--scale models; we do not claim SOTA or parity with large multilingual LLMs, and caution that even small adapted generators can be misused (spam/misinformation). We therefore emphasize research-only release practices and recommend additional safeguards for any deployment. We track tokens and follow compute-aware training to limit environmental impact, and we plan to study extension to 7B+ backbones and \emph{parallel-instruction} supervision to assess transfer and risk at LLM scale.

\section*{Acknowledgments}
This research was enabled in part by equipment and support provided by Mila - Quebec AI Institute and the Digital Research Alliance of Canada (\texttt{alliancecan.ca}). This research was supported in part by the Natural Sciences and Engineering Research Council (NSERC) of Canada. We are grateful for the support from IVADO and the Canada First Research Excellence Fund.

We also acknowledge the use of Stanford's Agentic AI Reviewer\footnote{https://paperreview.ai/} and OpenAI's ChatGPT 5.1-Pro to assist with reviewing and proofreading the final manuscript.

\bibliography{custom}

\appendix

\section{Base Backpack Model Configurations}
\label{app:base-configs}

\begin{table*}[h]
\centering
\small
\setlength{\tabcolsep}{6pt}
\begin{tabular}{lcccccc}
\toprule
\textbf{Model} & \textbf{Layers} & \textbf{Hidden} & \textbf{Heads} & \textbf{Senses} & \textbf{Positions} & \textbf{Params (M)} \\
\midrule
\textit{Small}  & 12 & 768  & 12 & 16 & 512 & 124 \\
\textit{Medium} & 24 & 1024 & 16 & 16 & 512 & 345 \\
\textit{Large}  & 36 & 1280 & 20 & 16 & 512 & 762 \\
\bottomrule
\end{tabular}
\caption{Configuration of base English Backpack models used for cross-lingual adaptation. Each model employs GELU activations, dropout of 0.1 across attention, residual, and embedding layers, and a sense intermediate scaling factor of 4.}
\label{tab:backpack-configs}
\end{table*}

The base English Backpack models were initialized with GPT-2 backbones of varying sizes, following the configurations summarized in Table~\ref{tab:backpack-configs}. 

All models use 16 sense embeddings per token and a shared vocabulary of 50,264 tokens. Hyperparameters such as dropout and layer normalization follow the original GPT-2 settings. Each variant was trained on OpenWebText prior to cross-lingual adaptation.

\section{Scheduler Details}
\label{app:scheduler}

Loss weights and InfoNCE temperatures are controlled by a cosine-based interpolation schedule parameterized by normalized training progress $p \in [0,1]$.  
Let $a$ denote the proportion of steps allocated to the alignment phase and $z$ the point at which the polish phase begins.  
Each weight $w$ and temperature $\tau$ is updated as a smooth function of $p$ according to  
\[
\text{interp}(a,b,t) = b + \tfrac{1}{2}(a-b)\bigl[1+\cos(\pi t)\bigr],t \in [0,1].
\]

\paragraph{Alignment phase ($p \le a$).}
\[
w_{\text{sns}}, w_{\text{ctx}} \text{ high}, \qquad w_{\text{lm}} \approx 0,
\]
with weights interpolated from their alignment values toward mid-phase values.  
Temperatures $\tau_{\text{sns}}$ and $\tau_{\text{ctx}}$ decay gradually from their initial settings.

\paragraph{Joint phase ($a < p < z$).}
\[
w_{\text{sns}}, w_{\text{ctx}} \text{ decreasing}, \qquad
w_{\text{lm}} \text{ increasing},
\]
following cosine interpolation between mid- and tail-phase targets.  
Temperatures continue to decrease until reaching their final values.

\paragraph{Polish phase ($p \ge z$).}
\[
w_{\text{sns}}, w_{\text{ctx}} \text{ low}, \qquad
w_{\text{lm}} \text{ high}.
\]
At this stage, training proceeds with high emphasis on the language modeling objective to improve fluency in the target language.

This schedule provides a smooth curriculum across objectives, avoiding discontinuities in optimization dynamics while emphasizing different learning signals at appropriate stages of adaptation.

We initially experiment with 20\%-50\%-30\% alignment/joint/polish phase boundaries, but have found that the LM loss still reduces considerably. We then shifted to a 20\%-30\%-50\% schedule and have kept this for all our experiments to maintain consistency.

\section{\textsc{SENSIA} Adaptation Setup}
\label{app:adaptation-setup}

We adapt the pretrained English Backpack model (\textsc{GPT-2 small} backbone) to each target language using parallel data from the OPUS corpus.
Training is performed for 150K steps with a learning rate of $5\times10^{-5}$, batch size of 64, warmup ratio of 0.1, gradient clipping at 1.0, and label smoothing of 0.05.
All runs use mixed-precision (FP16) optimization and the same random seed (1234).

Each adaptation follows the three-phase schedule described in Section~\ref{sec:training-objectives}: an \textit{alignment} phase (20\% of training steps), a \textit{joint} phase (30\%), and a \textit{polish} phase (50\%). Loss weights are progressively annealed between $(w_{\text{sns}}, w_{\text{ctx}}, w_{\text{lm}}) = (0.54, 0.44, 0.02)$ in the alignment phase, $(0.40, 0.40, 0.20)$ in the joint phase, and $(0.15, 0.15, 0.70)$ in the polish phase.
Both sense and context temperatures are fixed at $\tau_{\text{sns}} = 0.05$ and $\tau_{\text{ctx}} = 0.07$, and the sense pool temperature is 0.7.
During the polish phase, the sense vectors and sense-weighting network are frozen to prevent representational drift while the transformer and LM head continue to update.

Evaluation is conducted every 2K steps on the FLORES-200 dev set, tracking recall@1, perplexity, and sense-weight entropy.
All hyperparameters are held constant across Estonian, Indonesian, Swahili, and Turkish, differing only in dataset configuration (\texttt{eng-tgt}) and OPUS data splits.

\section{Dataset Construction}
\label{app:data-construction}

Parallel training data are built automatically using a unified pipeline that filters and merges sentence pairs from multiple OPUS corpora.  
For each target language, we run the same data-building script with language-specific corpus selections.  
The pipeline downloads OPUS data, removes duplicates, applies sentence-length and ratio filters, computes sentence similarity using LaBSE \cite{feng-etal-2022-language} embeddings, and outputs a final merged dataset capped at two million aligned pairs.  
The pipeline uses OpusTools\footnote{https://github.com/Helsinki-NLP/OpusTools} to retrieve data from online sources.

Sentence pairs are retained only if they satisfy a length ratio constraint (typically $0.6 < \frac{|x_{\text{src}}|}{|x_{\text{tgt}}|} < 1.7$) and exceed a LaBSE cosine similarity threshold ($\ge 0.80$--$0.82$ depending on the language).  
Each corpus contributes up to one million pairs before sampling to reach the target size of two million examples.  
The process automatically cleans temporary files after successful export and logs the filtering statistics for reproducibility.

Corpora used to build each language pair include:
\begin{itemize}
    \item \textbf{Estonian (en--et):} Tatoeba, WikiMatrix, CCAligned, OpenSubtitles, TED2020, GlobalVoices.
    \item \textbf{Indonesian (en--id):} JW300, WikiMatrix, TED2020, GlobalVoices, QED, CCAligned.
    \item \textbf{Swahili (en--sw):} JW300, Tatoeba, GlobalVoices, QED, WikiMatrix, CCAligned, TED2013, CCMatrix.
    \item \textbf{Turkish (en--tr):} SETimes, OpenSubtitles, WikiMatrix, TED2020, GlobalVoices, CCAligned.
\end{itemize}

All datasets are built with a batch size of 2048, a maximum of one to three million pairs per corpus before filtering, and final sampling to two million sentence pairs per language.  
Length ratio, similarity thresholds, and corpus composition are held constant across languages except for Swahili, which uses a wider length ratio range ($0.6$--$1.7$) and slightly lower similarity cutoff (0.80) to accommodate noisier data sources.

\section{Benchmarking Implementation Details}
\label{app:benchmarking}

\paragraph{Evaluation procedure.}
For all benchmarks, model predictions are scored using token-level log-likelihoods from the language modeling head.  
Given a context $C$ and a candidate continuation $A$, we compute its conditional probability  
$p(A \mid C) = \exp\!\left(\frac{1}{|A|}\sum_{t=1}^{|A|}\log p(a_t \mid a_{<t}, C)\right)$,  
where log-probabilities are averaged over all tokens.  

Following prior work, two complementary scoring schemes are used:  
\begin{enumerate}
    \item a conditional likelihood (\textit{cond}) method, which evaluates $p(A \mid C)$ directly, and
    \item a pointwise mutual information (PMI) formulation, $p(A \mid C) - p(A)$, which corrects for unconditional continuation bias.
\end{enumerate}

For each example, the model selects the candidate with the highest final score as the prediction.  
Accuracy is the proportion of correctly selected completions, and perplexity (PPL) is the exponentiated negative mean log-likelihood per token.

\paragraph{Scoring parameters.}
We perform a grid search for our models over interpolation weights $\lambda$ and $\alpha$ that combine PMI and conditional scores into a unified objective:
\[
\begin{aligned}
s(A \mid C) = (1 - \lambda)\,\mathrm{PMI}(A, C) + \\
\lambda\,\log p(A \mid C) + \alpha\,\mathrm{len}(A),
\end{aligned}
\]
where the final term penalizes length bias in generation.  
Grid search was performed on the XCOPA validation set over $\lambda \in \{0.0, 0.1, \ldots, 1.0\}$ and $\alpha \in \{0.0, 0.1, \ldots, 1.0\}$, selecting parameters that maximized validation accuracy per language and task.  
The optimal parameters used for all reported results are summarized in Table \ref{tab:lambda_alpha}.

\begin{table}[t]
\centering
\small
\setlength{\tabcolsep}{5pt}
\begin{tabular}{lcccccc}
\toprule
\multirow{2}{*}{\textbf{Language}} & \multicolumn{2}{c}{\textbf{XCOPA}} & \multicolumn{2}{c}{\textbf{XStoryCloze}} & \multicolumn{2}{c}{\textbf{Belebele}} \\
\cmidrule(lr){2-3} \cmidrule(lr){4-5} \cmidrule(lr){6-7}
 & $\boldsymbol{\lambda}$ & $\boldsymbol{\alpha}$ & $\boldsymbol{\lambda}$ & $\boldsymbol{\alpha}$ & $\boldsymbol{\lambda}$ & $\boldsymbol{\alpha}$ \\
\midrule
Estonian  & \textit{cond} & 0.10 & --- & --- & 1.00 & 0.20 \\
Indonesian& 0.50 & 0.10 & \textit{cond} & 1.00 & 1.00 & 0.20 \\
Swahili   & 0.60 & 0.70 & \textit{cond} & 0.70 & 0.90 & 0.90 \\
Turkish   & 0.40 & 0.40 & --- & --- & 0.40 & 0.50 \\
\bottomrule
\end{tabular}
\caption{Optimal $\lambda$ and $\alpha$ parameters per language and task. ``\textit{cond}'' indicates the use of conditional scoring only (no PMI term).}
\label{tab:lambda_alpha}
\end{table}

\paragraph{Baseline results.}
Scores for larger multilingual baselines and previously published models are sourced directly from prior work for transparency and consistency:
\begin{itemize}
    \item \textbf{XCOPA:} BLOOMZ-7B1 and XGLM-7B5 from the BLOOMZ paper; Goldfish from the Goldfish paper.
    \item \textbf{XStoryCloze:} BLOOMZ-7B1 and XGLM-7B5 from the BLOOMZ paper; Goldfish from the Goldfish paper.
    \item \textbf{Belebele:} XGLM-7B5 and Goldfish from the Goldfish paper.  
    BLOOMZ-7B1 results were obtained by running the released Goldfish evaluation scripts on BLOOMZ to ensure consistent scoring across models.
\end{itemize}

We obtain GPT-2 + FT and MCL scores via our own implementation of these methods. Both GPT-2 + FT and MCL use the same data and token budget as \textsc{SENSIA}-adapted models during training to keep comparisons data-matched. During decoding, we also use the same (optimal) hyperparameters used by the \textsc{SENSIA}-adapted models to ensure comparability.

\paragraph{Reproducibility.}
All \textsc{SENSIA} evaluations were run with fixed random seeds and deterministic decoding order.  
Scoring outputs, predictions, and selected hyperparameters are included with the released evaluation materials to facilitate replication.

\section{Generation Examples}
\label{app:generation}

We test with short translated prompts (for example, \textit{``President Barack Obama said''}) to qualitatively assess fluency.  
The adapted model produces fluent, well-structured continuations in all four target languages:
\begin{itemize}
    \item \textbf{Estonian:} President Barack Obama ütles, et ta on kõige kallim, kes on kunagi olnud.\\
    \textit{``President Barack Obama said he is the most expensive person who has ever lived.''}
    \item \textbf{Indonesian:} Presiden Barack Obama mengatakan bahwa kami tidak akan mengambil keuntungan dari kemampuan kami untuk mengumpulkan informasi...\\
    \textit{``President Barack Obama said that we will not take advantage of our ability to collect information...''}
    \item \textbf{Swahili:} Rais Barack Obama alisema kuwa Marekani inaamini kuwepo kwa makubaliano ya Paris...\\
    \textit{``President Barack Obama said the United States believes in the Paris Agreement on climate change.''}
    \item \textbf{Turkish:} Başkan Barack Obama şöyle dedi: Bunu yapmak için çok çalıştım.\\
    \textit{``President Barack Obama said: I worked very hard to do this.''}
\end{itemize}

\section{Additional Analysis and Ablation Details}
\label{app:extra-ablations}

In this section we provide the exact formulations and practical
protocols for the analyses and empirical ablations in
Sections~\ref{sec:sense-ablations} and~\ref{sec:empirical-ablations}.
We follow the notation from the main methodology section:
each token is represented by a mixture of $K$ latent sense vectors
$s_{t,1},\dots,s_{t,K}$ with mixture weights $\pi_{t,k}$, and
sentence-level vectors are $\ell_2$-normalized before contrastive
training.

\subsection{Sense-Space Analyses (Section~\ref{sec:sense-ablations})}

\paragraph{Data and Languages.}
All analyses use the FLORES-200 devtest split~\citep{guzman2019flores, costa-jussa-etal-2022-nllb}, which provides balanced English--target pairs for each evaluation language: Indonesian (ind), Swahili (swh), Estonian (est), and Turkish (tur).  
Each English--target pair was drawn from the same dataset used for tracking recall@1 and perplexity during training, ensuring consistency between intrinsic and extrinsic evaluations.

\paragraph{Word Alignment.}
Word-level alignments were obtained using \textsc{SimAlign}~\citep{jalili-sabet-etal-2020-simalign} with the multilingual \textsc{MiniLM} encoder in \textit{itermax} mode, which iteratively refines bidirectional attention for high-precision alignments.  
We retained only one-to-one alignments to ensure semantic comparability between tokens.  
Multiword expressions and one-to-many alignments were excluded, as were rare or function words that did not appear in both languages.

\paragraph{Sense Extraction.}
For each aligned word pair $(w_{\text{en}}, w_{\text{tgt}})$, we extracted the model’s pre-context sense embeddings from the Backpack sense module.
When a target-language word was tokenized into multiple subwords under the shared BPE vocabulary, we pooled the corresponding sense vectors by averaging their activations before normalization.  
To ensure stable estimates, we filtered to include only content words (nouns, verbs, adjectives) appearing at least five times in the corpus.  
All sense vectors were $\ell_2$-normalized and mean-centered before computing correlations or alignments.

\paragraph{Sense topology alignment.}
For each aligned word pair $(w_{\mathrm{src}}, w_{\mathrm{tgt}})$ from
FLORES-200 obtained with SIMALIGN, we extract the \emph{pre-context}
sense vectors from the Backpack sense module.
This yields matrices
$S_{\mathrm{src}}, S_{\mathrm{tgt}} \in \mathbb{R}^{K \times d}$,
where each row corresponds to one of the $K$ senses of the word in the
given language. Because a target-language word may span multiple GPT-2
BPE tokens, we first pool sense vectors over all subword positions
belonging to the word span (simple average over positions), obtaining
$V_{\mathrm{src}}, V_{\mathrm{tgt}} \in \mathbb{R}^{K \times d}$.%
\footnote{In practice, we discard (word, language) pairs where fewer
than $K$ senses are available after pooling, which is rare under our
Backpack configuration.}

We then mean-center and row-normalize the sense matrices:
\begin{equation}
  \tilde{V}
  = \mathrm{normalize}
    \Bigl(
      V - \frac{1}{K} \sum_{k=1}^K V_k
    \Bigr),
\end{equation}
where $V_k$ is the $k$-th row of $V$ and
$\mathrm{normalize}(\cdot)$ applies row-wise $\ell_2$-normalization.
All sense vectors are thus mean-centered and unit-norm before we
compare geometries.%

The local geometry for a word in each language is summarized by a
cosine Gram matrix
\begin{equation}
  G_{\mathrm{src}} = \tilde{V}_{\mathrm{src}} \tilde{V}_{\mathrm{src}}^\top,
  \qquad
  G_{\mathrm{tgt}} = \tilde{V}_{\mathrm{tgt}} \tilde{V}_{\mathrm{tgt}}^\top.
\end{equation}
We flatten the upper-triangular entries (excluding the diagonal) of
each Gram matrix into vectors
$g_{\mathrm{src}}, g_{\mathrm{tgt}} \in
\mathbb{R}^{K(K-1)/2}$, and compute the
Spearman rank correlation
\begin{equation}
  \rho(w_{\mathrm{src}}, w_{\mathrm{tgt}})
  = \mathrm{Spearman}\!\left(
      g_{\mathrm{src}}, g_{\mathrm{tgt}}
    \right).
\end{equation}
This $\rho$ score measures how similarly the two languages organize
the $K$ senses of the aligned word pair.%
We report the mean $\rho$ over all sampled aligned word pairs per
language, together with the corresponding control model (LM-only
finetuning with $w_{\text{sns}} = w_{\text{ctx}} = 0$,
$w_{\text{lm}} = 1.0$).

For statistical significance, we run $10{,}000$ paired bootstrap
iterations, resampling aligned word pairs with replacement at each
iteration and recomputing the average $\rho$ for both the adapted model
and the control. The reported $\rho$ values in Table~\ref{tab:sense_topology_results}
are averages over these iterations, and the differences
$\Delta \rho$ are significant at the 95\% level
($p < 0.01$).%

\paragraph{Orthogonal Procrustes analysis.}
For the global structure analysis, we focus on
\emph{mixture-weighted} sense embeddings. For each aligned word pair,
we compute sense mixtures using the model's sense-weighting network:
given sense vectors $s_{t,1},\dots,s_{t,K}$ and per-sense mixture
weights $\pi_{t,1},\dots,\pi_{t,K}$ for the word,
\begin{equation}
  e = \sum_{k=1}^K \pi_{t,k}\, s_{t,k},
\end{equation}
followed by $\ell_2$-normalization of $e$.
We collect $N = 10{,}000$ aligned word pairs per language, forming
matrices
$E, T \in \mathbb{R}^{N \times d}$ whose rows are the English and
target-language mixture embeddings, respectively.%

We then solve the orthogonal Procrustes problem
\begin{equation}
  Q^\star
  =
  \arg\min_{Q : Q^\top Q = I}
  \;\bigl\| T Q - E \bigr\|_F,
\end{equation}
whose solution is
$Q^\star = U V^\top$,
where $T^\top E = U \Sigma V^\top$ is the singular value decomposition
(SVD). After alignment, we compute the average cosine similarity
\begin{equation}
  \mathrm{Cosine}
  =
  \frac{1}{N}
  \sum_{i=1}^N
  \cos\bigl(T_i Q^\star, E_i\bigr),
\end{equation}
where $T_i$ and $E_i$ are the $i$-th rows of $T$ and $E$, respectively.
We average this score over five random subsamples of $N=10{,}000$
aligned pairs to ensure stability.
Table~\ref{tab:procrustes_results} reports the mean cosine for both the adapted
model and the LM-only control, along with the difference
$\Delta \cos$.

\subsection{Empirical Ablation Protocols (Section~\ref{sec:empirical-ablations})}

Unless otherwise noted, all ablations reuse the \textsc{SENSIA} adaptation
setup from Appendix \ref{app:adaptation-setup} (learning rate, batch size, phase schedule, loss
weights, and temperatures), and the evaluation protocol from
Appendix \ref{app:benchmarking} (conditional/PMI scoring and length normalization).%
Each ablation is either an inference-time modification of a fixed
checkpoint or a separate training run with a single component turned
off, and all scores are reported on the same FLORES-200 devtest split
or downstream benchmarks as in the main results.

\paragraph{Sense-mixture ablation.}

We test whether downstream performance depends on the sense-mixture
mechanism by modifying the mixture weights $\pi_{t,k}$ at inference
time while keeping all model parameters fixed.

For each token $t$ and sense index $k$, let $\pi_{t,k}$ be the
mixture weight produced by the sense-weight network, and let
$\mathrm{Top}\text{-}k_t$ be the set of indices for the $k$ most
probable senses at position $t$. We compare:

\begin{enumerate}
  \item \textbf{Full mixture} (default \textsc{SENSIA}): use the learned
  $\pi_{t,k}$ unchanged.

  \item \textbf{Top-$k$ mixture} (we use $k=1$):
  \begin{equation}
    \pi'_{t,k} =
    \begin{cases}
      \displaystyle
      \frac{\pi_{t,k}}{\sum_{j \in \mathrm{Top}\text{-}k_t} \pi_{t,j}}
      & k \in \mathrm{Top}\text{-}k_t, \\[8pt]
      0 & \text{otherwise.}
    \end{cases}
  \end{equation}

  \item \textbf{Uniform mixture}:
  \begin{equation}
    \pi^{\mathrm{uni}}_{t,k} = \frac{1}{K}
    \qquad \forall\,k \in \{1,\dots,K\}.
  \end{equation}
\end{enumerate}

In all three cases, the token representation is recomputed as
$h_t = \sum_k \pi_{t,k} s_{t,k}$ (or with $\pi'_{t,k}$ /
$\pi^{\mathrm{uni}}_{t,k}$), and the resulting sequence is fed through
the frozen transformer and LM head.

We evaluate each variant on the FLORES-200 devtest set, computing
average token-level cross-entropy
\begin{equation}
  \mathrm{CE}
  =
  -\frac{1}{T}
  \sum_{t=1}^T
  \log p_\theta(y_t \mid y_{<t}, x),
\end{equation}
where $T$ is the number of target tokens and $x$ is the source
sentence.
Table~\ref{tab:sense_mixture_ablation} reports CE for four target languages, all
evaluated with the same adapted small Backpack checkpoint.

\paragraph{Loss and phase ablations.}

We ablate individual loss terms and training phases using Indonesian as
the target language.
Recall the total objective
\begin{equation}
  \mathcal{L}_{\text{total}}
  = w_{\text{sns}} \mathcal{L}_{\text{sns}}
  + w_{\text{ctx}} \mathcal{L}_{\text{ctx}}
  + w_{\text{lm}} \mathcal{L}_{\text{lm}},
\end{equation}
with phase-dependent weights
$(w_{\text{sns}}, w_{\text{ctx}}, w_{\text{lm}})$ annealed according
to the schedule in Appendix~C.
For each ablation variant in Table~\ref{tab:loss_phase_ablation}, we train a
separate model with the following modifications:

\begin{itemize}
  \item \textbf{No Context loss:}
  set $w_{\text{ctx}} = 0$ at all steps; keep
  $w_{\text{sns}}$ and $w_{\text{lm}}$ as in the default schedule.

  \item \textbf{No Sense loss:}
  set $w_{\text{sns}} = 0$ at all steps; keep
  $w_{\text{ctx}}$ and $w_{\text{lm}}$ as in the default schedule.

  \item \textbf{No LM loss:}
  set $w_{\text{lm}} = 0$ at all steps; keep
  $w_{\text{sns}}$ and $w_{\text{ctx}}$ unchanged.

  \item \textbf{No Alignment phase:}
  skip the alignment phase entirely and start from the joint-phase
  weights, then proceed to the polish phase as usual.

  \item \textbf{No Joint phase:}
  run the alignment phase and then jump directly to the polish phase.

  \item \textbf{No Polish phase:}
  run only the alignment and joint phases (truncating total training
  steps to match the baseline), with the same per-phase weights.
\end{itemize}

All other hyperparameters (learning rate, batch size, temperatures,
phase boundaries, etc.) follow Appendix~C, and we reuse the same
evaluation protocol and scoring parameters as the full model.
XCOPA, XStoryCloze, Belebele, and FLORES perplexity are evaluated on
the same splits as Table~\ref{tab:reasoning_comprehension}, producing the numbers
in Table~\ref{tab:loss_phase_ablation}.%

\paragraph{Scaling experiments.}

For scaling, we extend Backpack to GPT-2 medium (345M) and large
(762M) backbones. Pretraining follows the OpenWebText setup of
\citet{hewitt2023backpack}, with token budgets chosen according to the
Chinchilla scaling law to balance model size and data scale.%
We then adapt each backbone to Indonesian and Swahili using the same
\textsc{SENSIA} schedule as for the small model.

Generation and scoring hyperparameters (cond/PMI interpolation and
length normalization) are held fixed across model sizes to isolate the
effect of scale, giving the results in Table~\ref{tab:scaling-results}.

\paragraph{Low-resource adaptation.}

To approximate low-resource conditions, we downsample the Indonesian
and Swahili parallel corpora used in the main experiments to a
subset of 100k sentence pairs per language.%
Downsampling is performed by uniformly sampling without replacement
from the full OPUS-derived corpus (after all filtering), and we keep
the Backpack architecture, \textsc{SENSIA} schedule, loss weights, and
hyperparameters identical to the full-data setting.

Each low-resource model is trained from the same English Backpack
initialization and evaluated on the same XCOPA, XStoryCloze, and
Belebele test splits as the full-data model. We reuse the same
decoding/scoring setup (Appendix \ref{app:benchmarking}), so differences in accuracy and
FLORES perplexity in Table~\ref{tab:low-resource} are attributable
solely to the reduction in parallel data.

\paragraph{Non-Latin script stress test.}

For the Chinese stress test, we adapt the same English Backpack model
to Chinese using English--Chinese parallel data mined from OPUS, but
\emph{keep the original English GPT-2 BPE tokenizer unchanged}.%
Chinese text is thus represented as long sequences of over-segmented
subword fragments, intentionally placing the model in a hostile
tokenization regime.

We reuse the adaptation hyperparameters, phase schedule, and loss
weights from the main experiments, and evaluate on Chinese XCOPA,
Chinese XStoryCloze, and the Chinese split of Belebele, comparing
directly against the monolingual Chinese Goldfish baseline under the
same evaluation protocol and scoring setup.%
Table~\ref{tab:chinese-test} reports per-task accuracies for this adversarial
setting and supports our conclusion that sense
adaptation cannot fully overcome a severely mismatched tokenizer,
even though it still yields modest gains on some narrative-style tasks.

\end{document}